\documentclass{article}
\usepackage{amsmath}
\usepackage{amssymb}
\usepackage{microtype}
\usepackage{graphicx}
\usepackage{subfigure}
\usepackage{booktabs} 

\usepackage{hyperref}
\usepackage{array}

\usepackage[accepted]{icml2021}

\icmltitlerunning{Image Animation with Keypoint Mask}

\begin{document}

\twocolumn[
\icmltitle{Image Animation with Keypoint Mask}

\icmlsetsymbol{equal}{*}

\begin{icmlauthorlist}
\icmlauthor{Or Toledano}{tlv}
\icmlauthor{Yanir Marmor}{tlv}
\icmlauthor{Dov Gertz}{tlv}

\end{icmlauthorlist}

\icmlaffiliation{tlv}{Tel Aviv University}

\icmlcorrespondingauthor{Or Toledano}{ortoledano@protonmail.com}
\icmlcorrespondingauthor{Yanir Marmor}{yanirmr@gmail.com}
\icmlcorrespondingauthor{Dov Gertz}{dovgertz1@gmail.com}

\icmlkeywords{Machine Learning, ICML}

\vskip 0.3in
]
\printAffiliationsAndNotice{}
\begin{table}[t]
\caption{\textbf{Motion transfer example:} Given a YouTube clip of a
Thai-Chi artist (top), and an image of another one, our method
transfers the one’s performance onto the other (bottom).}
\label{table:intro}
\vskip 0.15in
\begin{center}
\begin{small}
\begin{sc}
\begin{tabular}{m{1.0cm}m{1.0cm}m{1.0cm}m{1.0cm}m{1.0cm}m{1.0cm}}
\toprule
\includegraphics[width=1cm, height=1cm]{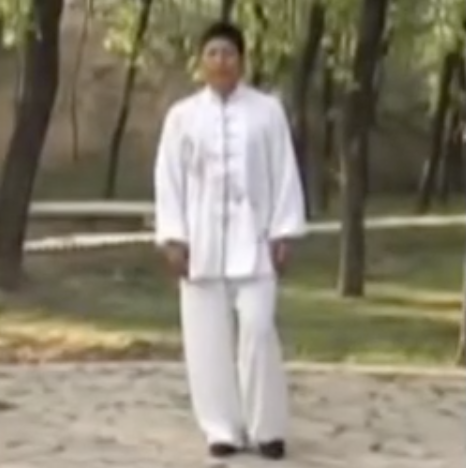} &
\includegraphics[width=1cm, height=1cm]{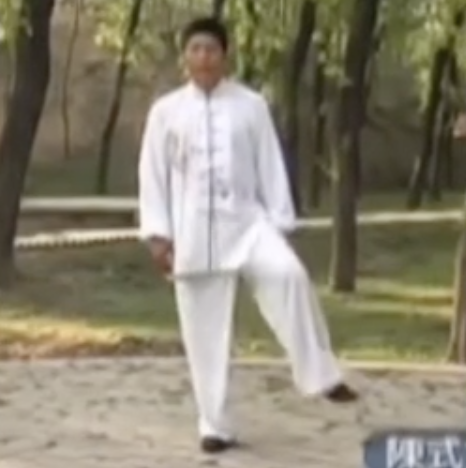} &
\includegraphics[width=1cm, height=1cm]{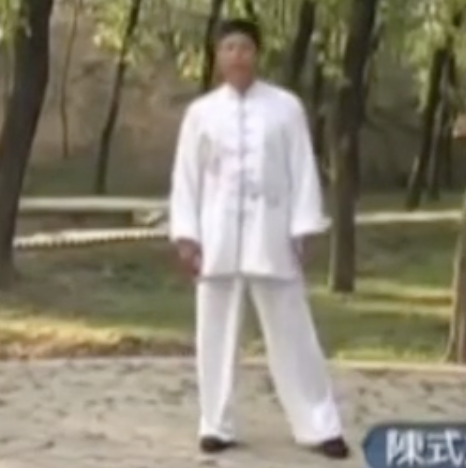} &
\includegraphics[width=1cm, height=1cm]{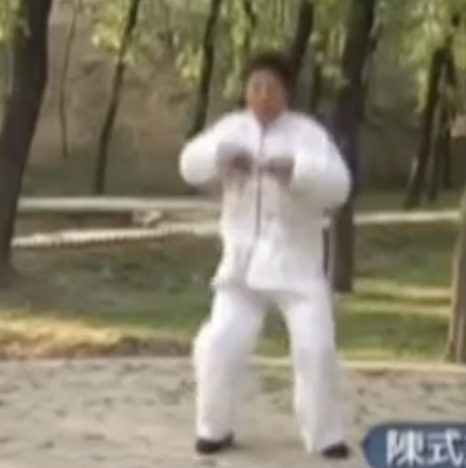} &
\includegraphics[width=1cm, height=1cm]{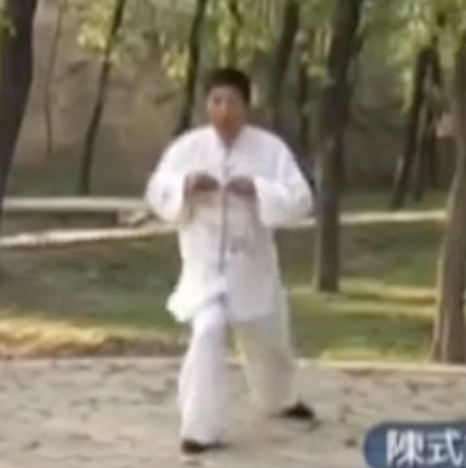} &
\includegraphics[width=1cm, height=1cm]{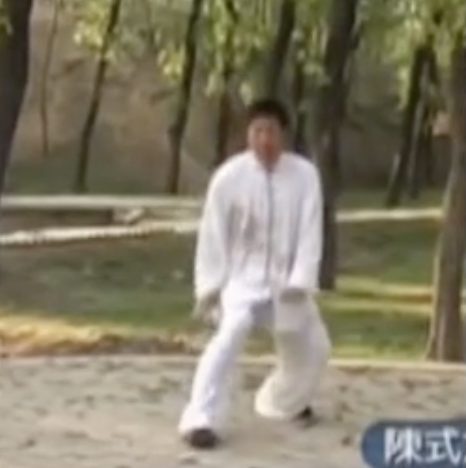}\\
\includegraphics[width=1cm, height=1cm]{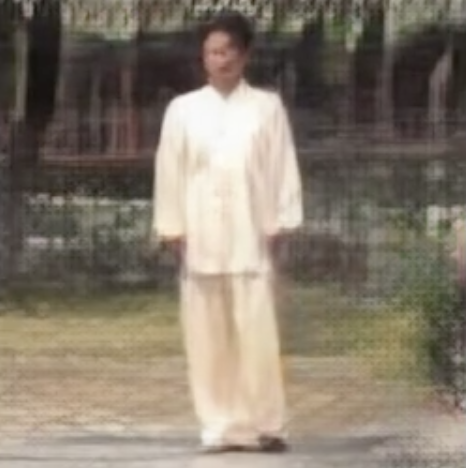} &
\includegraphics[width=1cm, height=1cm]{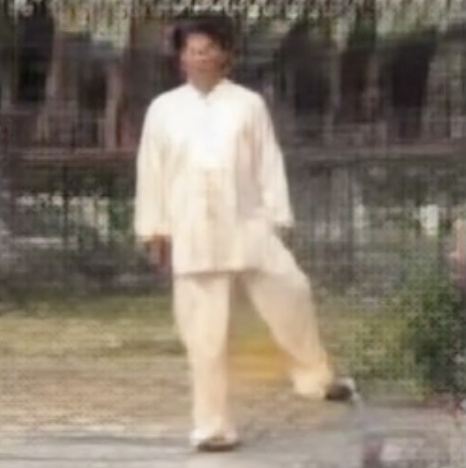} &
\includegraphics[width=1cm, height=1cm]{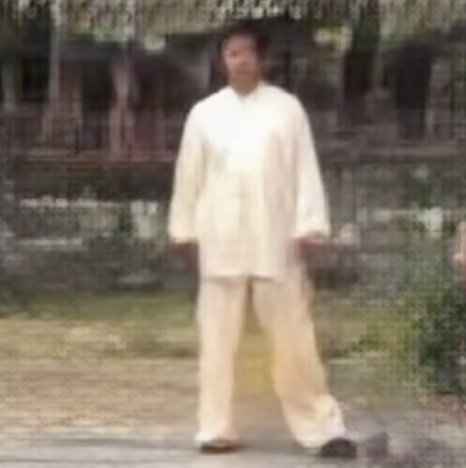} &
\includegraphics[width=1cm, height=1cm]{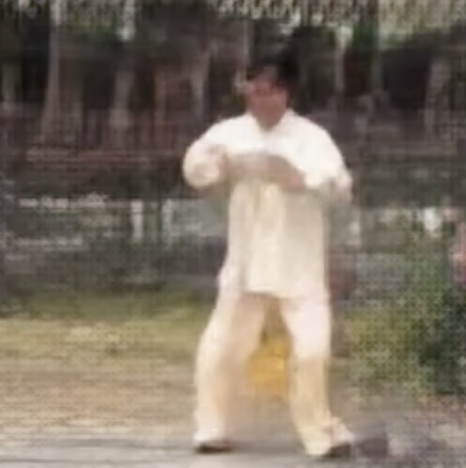} &
\includegraphics[width=1cm, height=1cm]{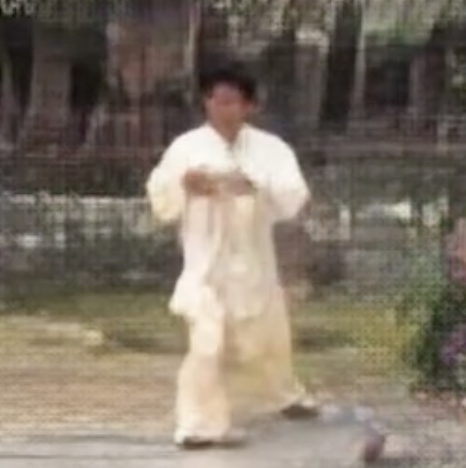} &
\includegraphics[width=1cm, height=1cm]{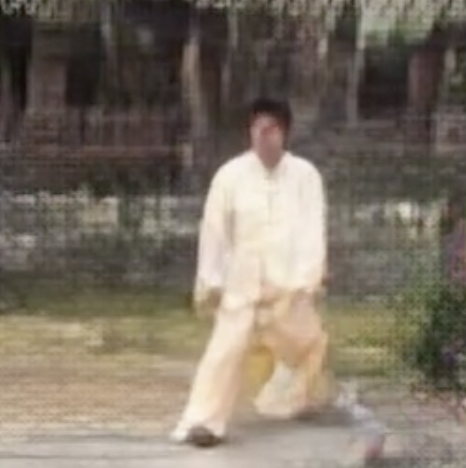}\\
\bottomrule
\end{tabular}
\end{sc}
\end{small}
\end{center}
\vskip -0.1in
\end{table}

\begin{abstract}
Motion transfer is the task of synthesizing future video frames of a single
source image according to the motion from a given driving video.
In order to solve it, we face the challenging complexity of motion representation
and the unknown relations between the driving video and the source image.
Despite its difficulty, this problem attracted great interests from
researches at the recent years, with gradual improvements. The goal is often
thought as the decoupling of motion and appearance, which is
may be solved by extracting the motion from keypoint movement.
We chose to tackle the generic, unsupervised setting, where we need to apply
animation to any arbitrary object, without any domain specific model for the
structure of the input.
In this work, we extract the structure from a keypoint heatmap, without an
explicit motion representation. Then, the structures from the image and the
video are extracted to warp the image according to the video, by a deep
generator. We suggest two variants of the structure from different steps in
the keypoint module, and show superior qualitative pose and quantitative
scores.
\end{abstract}

\section{Introduction}
Take a look at Figure ~\ref{table:intro} for two video sequences. The input
is a YouTube clip of a Thai-Chi artist (the driving subject) performing a
series of complex motions at the top row. Our algorithm's output is shown
at the bottom row. It refers to frames that appear to show a different person
(the source subject) executing the same motions. The crux is that the
source individual has never performed the same exact
sequence of motions as the driver video - they were photographed doing another movement
with no clear reference to the driving's actions.
By observing the figure, we can see that the source and the driving are of different
appearances, have different backgrounds, and are dressed differently - a
totally different style that we need to ignore when we extract the
structure.

In this study, we propose a simple, however surprisingly efficient approach
to the general motion transfer problem, which can be
applied to any domain, similarly to \cite{siarohin2019animating},
\cite{siarohin2020order}:
Given a source image of a person with the wanted appearance,
and a driving video with the wanted structure/geometry,
we synthesize a video by applying a deep motion generator per-frame. We
assist the generator by feeding it with compact structural representations
of the source and each frame of the driving video. That way, we obtain the
target of a video of the source appearance and the driving motion, given
any domain. Keeping the generator as a frame processor enables it to be
simple and fast to train and evaluate, but note that there are works which
take into account temporal information, such as
\cite{villegas2017learning} which make a use of LSTM.

Researchers of some notable works (Section~\ref{related}) observe that
keypoint-based pose preserves motion signatures over time, while
abstracting subject identities. We therefore use the keypoint-based pattern
without any other motion priors. We obtain the keypoints and use them to
animate images in a way that doesn't need any external information about the
subject or any assumption or prior about the scene.

Our contribution is twofold: first, we demonstrate that removing the
explicit motion prior from works such as FOMM is feasible to the task of
image animation, and creates a compact, faster to train and evaluate model;
And second, we make progress to achieve a better structure during
animation, although further work (Section~\ref{future}) is required to
refine our model.
We also split our approach by the proposal of
two methods for absolute and relative motion transfer
(Section~\ref{circles-mask}), which covers both needs (when the source isn't
aligned well with the driving and needs to be matched to the current
driving explicitly,
and when it is aligned and it is wanted and more natural
to only add the relative difference
between the current driving and the driving during the first frame).

\medskip

\section{Related Work}
\label{related}
Motion transfer has gained a lot of coverage over the last twenty years.
Early approaches based on manipulating existing video footage to generate
new content. These included searching for frames in which the body position
corresponds to a desired motion and using them to generate a new
content \cite{bregler1997video}. Our approach is equally designed for videos,
but rather than manipulating existing images, we learn to synthesize new
movements that were never seen before with the new identities.

A number of techniques are based on calibrated multi camera systems to scan
a target player and use an adapted 3D model of the target to control their
motion in a new frame \cite{cheung2004markerless}. Our solution instead
examines the transition of movement between 2D video subjects and refrains
from using data calibration, 3D space information or any other
domain specific prior.

Latest approaches concentrate on the disentanglement of appearance and
activity and synthesizing of new motion videos \cite{tulyakov2018mocogan}.
Similarly, we apply our representation of motion to different target
subjects to generate new motions. However, In contrast to these works,
we did not use GAN but an encoder-decoder approach.

Contrary to image animation approaches that were
common up until recently, keypoint-based methods are now thought to have
the ability to achieve high performance in the field of video reanimation.
Some notable works in
this area are \cite{siarohin2020order},
\cite{siarohin2019animating}, \cite{kim2019unsupervised},
\cite{balakrishnan2018synthesizing}, \cite{ma2017pose}
\cite{chan2019everybody}.

Our work does not depend on a strong
motion prior directly, but rather on a
structure mask derived from a keypoint detector of a motion-based model,
such as \cite{siarohin2020order}. The idea of using drawn keypoints as a
geometry representation (structural mask) was already used in image-to-image
translation works such as TransGaGa \cite{wu2019transgaga}, in addition to
some of the video reanimation works mentioned earlier.

The concept of using a structural mask in the context of image animation is
demonstrated in \cite{shalev2020image}. However, the current work differs
by basing the mask off a motion related keypoint module.
By doing so, we create a bottleneck for the network which is dependent on
the keypoint bottleneck used when training the keypoint module, in order to
achieve generalization. In addition, it simplifies the network, makes it
modular to the mask, and saves us the hassle of perturbing the input
hoping to achieve an identity-less mask.

We purpose an heatmap mask (Section~\ref{heatmap-mask}) for absolute
animation,
which differs from the drawn keypoint masks mentioned in previous works, in
addition to a classical keypoint mask (Section~\ref{circles-mask}) for
relative animation w.r.t. difference between the current and first frames in
the driving video.

\section{Methodology}
The network can be divided into two parts: obtaining the mask, and generating
the synthesized frame. The generator architecture is constant amongst all of
our variations, and can be described as low resolution generation from the
source image, source mask and driving mask; Followed by up-scaling of the
low scale synthesized prediction by passing it with the source frame to a
high resolution generator.
We created two different versions for the mask generator: the first,
heatmap based mask, which is limited to absolute motion transfer;
the second, keypoint based mask, which enables an optional relative motion
transfer.
The performance drop for our second version was expected because
its mask contained less structural information.
The first version is referred as a "keypoint heatmap mask" while the second
is referred as a "circles mask" or "keypoints after softmax" mask.

\subsection{First mask version for absolute motion transfer (with warping)}
\label{heatmap-mask}
\begin{figure}[ht]
\vskip 0.2in
\begin{center}
\centerline{\includegraphics[width=\columnwidth]{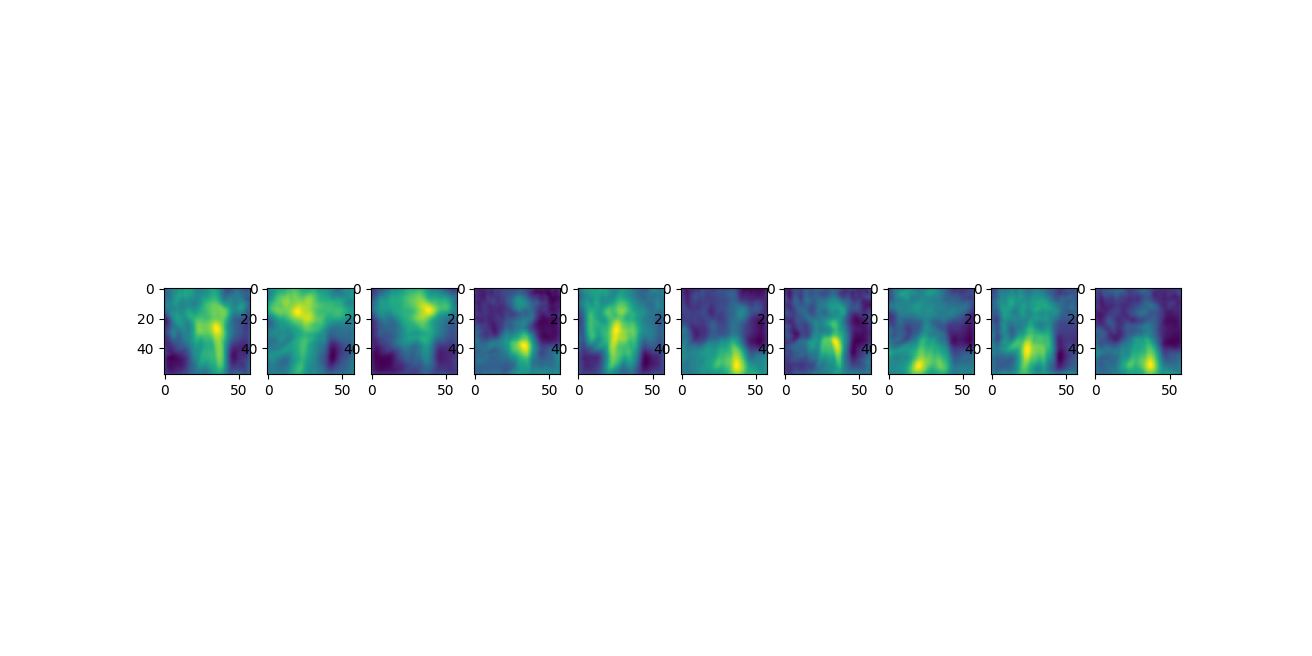}}
\caption{
$K$ channels of the keypoint detector network used in
\cite{siarohin2020order}, before the softmax activation. Our main motion
prior in this project.
}
\label{mask-10kp}
\end{center}
\vskip -0.2in
\end{figure}

Our main mask for the project is obtained from carefully observing the
keypoint module from \cite{siarohin2020order}. U-Net based keypoint
modules of this form, work by extracting features, which pass through a
$\textit{conv}$ layer to form a heatmap image $K$ channels, where $K$ is the
number of keypoints. Then, a softmax is performed over the channels, and
keypoints are extracted from the mean location of each heatmap channel.
By carefully debugging the code, and the expectation for a segmentation map
out of U-Net, we obtain Figure~\ref{mask-10kp}.

\begin{figure}[ht]
\vskip 0.2in
\begin{center}
\centerline{\includegraphics[width=\columnwidth]{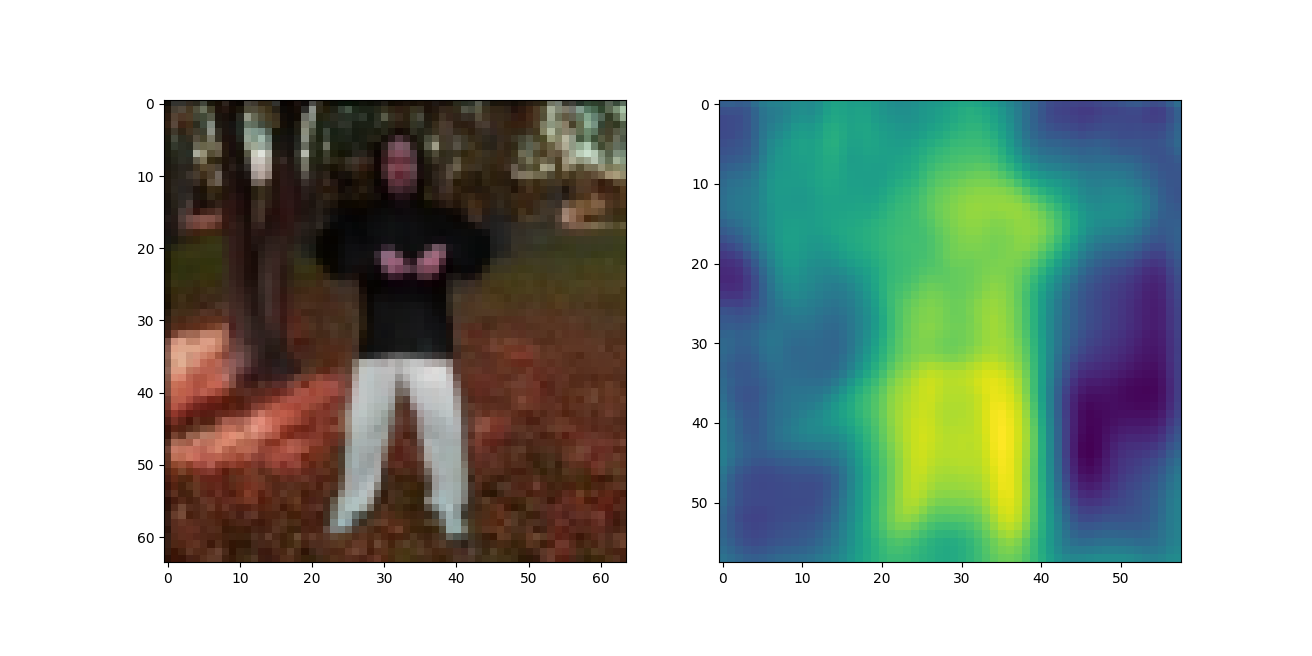}}
\caption{
The sum of the $K$ channels which is fed as a structural mask into the
generator.
}
\label{mask-sum}
\end{center}
\vskip -0.2in
\end{figure}

Summing over the channels, we get Figure~\ref{mask-sum}.

which is our output mask, aka the heatmap, pre softmax mask.

\subsection{Second mask version for relative motion transfer}
\label{circles-mask}
We purposes an additional "circles" only mask
which can be used in the context
of relative motion transfer during animation, as in
\cite{siarohin2020order}, which isn't possible with the previous heatmap mask.
The mask captures the image's geometry representation \cite{wu2019transgaga},
and by requiring it to be represented as keypoints with a center, we can use
the relative coordinates for the animation. This is done by replacing the
driving keypoints with the source keypoints moved according to the driving
displacement from the first frame, during animation. This module, though,
did not do as well as our heatmap mask module in the video reconstruction
task (Table~\ref{table:results}), but did generalize and managed to
preserve the background during animation, due to to the
small mask (Table~\ref{table:images}).

While relative motion transition is not always desired, this work shows that
a keypoint-only-prior-based module is feasible for the task.
Since the only information contained in the pair of masks is the keypoint
displacement, our deep network can only attempt to approximate a zero order
approximation, we can anticipate results that are more close to
\cite{siarohin2019animating}.

\begin{figure}[ht]
\vskip 0.2in
\begin{center}
\centerline{\includegraphics[width=\columnwidth]{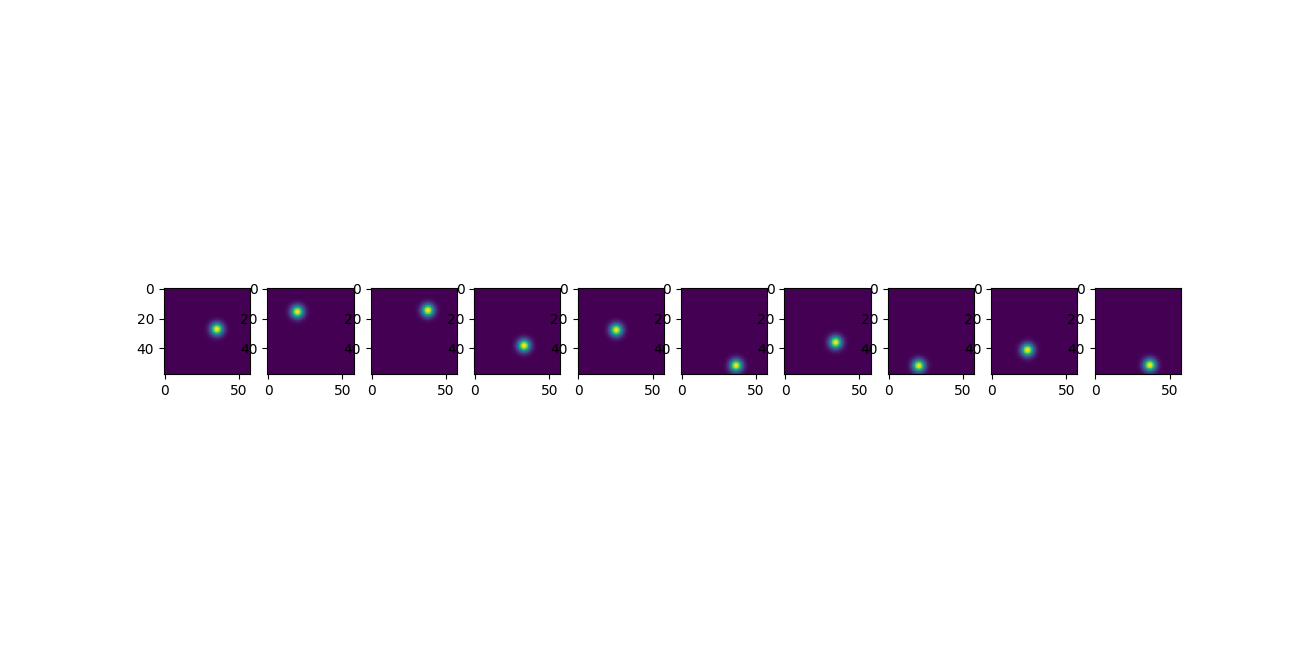}}
\caption{
$K$ channels of the keypoint detector network used in
\cite{siarohin2020order}, after the softmax activation and Gaussian fit.
}
\label{softmax-10kp}
\end{center}
\vskip -0.2in
\end{figure}

By taking the softmax over the heatmap, we get Figure~\ref{softmax-10kp}.

\begin{figure}[ht]
\vskip 0.2in
\begin{center}
\centerline{\includegraphics[width=\columnwidth]{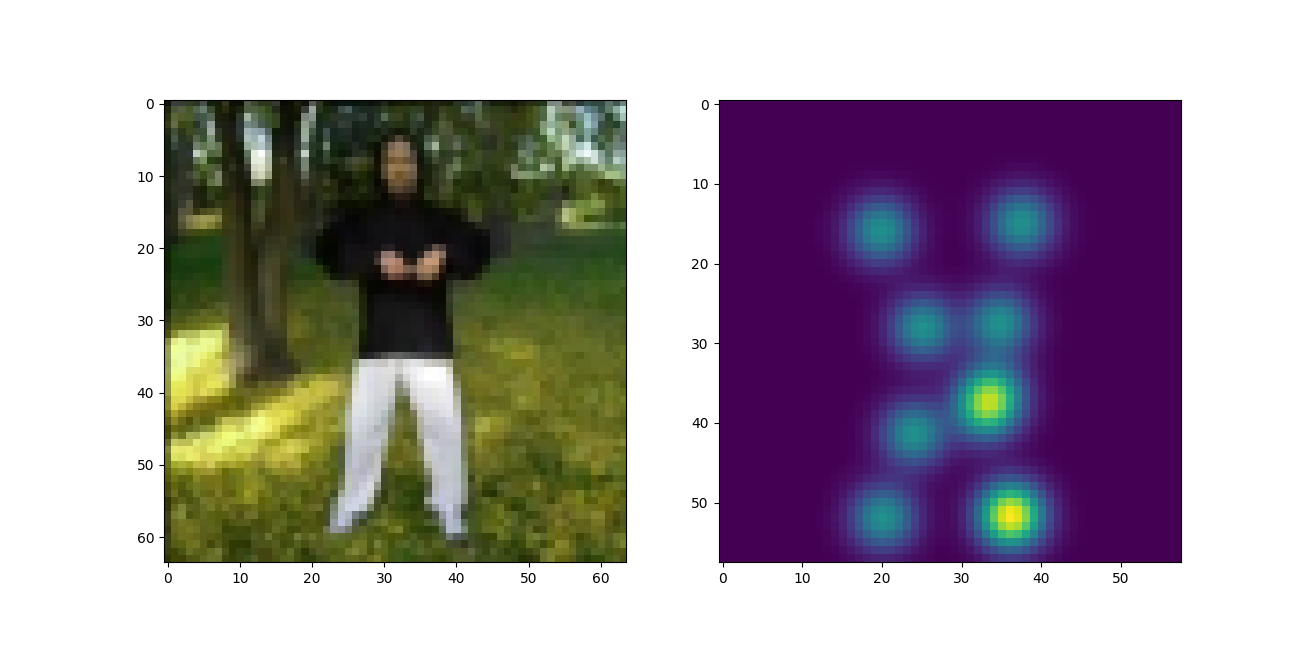}}
\caption{
The sum of the $K$ channels which is fed as a structural mask into the
generator.
}
\label{softmax-sum}
\end{center}
\vskip -0.2in
\end{figure}

Summing over the channels, we get Figure~\ref{softmax-sum}.

which is our output mask, aka the keypoints after softmax mask.

\subsection{Architecture}
\label{method}

\begin{figure}[ht]
\vskip 0.2in
\begin{center}
\centerline{\includegraphics[width=\columnwidth]{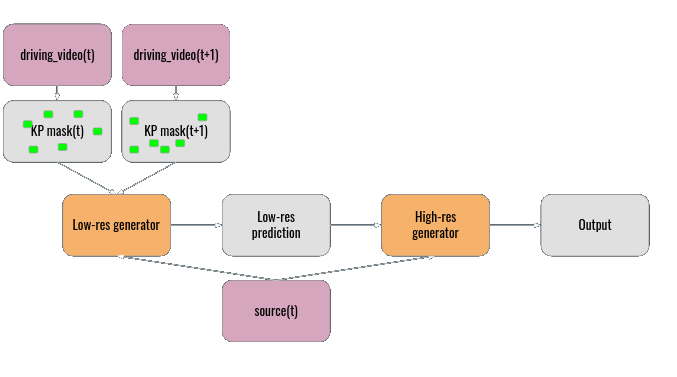}}
\caption{Architecture of the model. The keypoint mask can be either a
keypoint heatmap such as in Figure~\ref{mask-sum}, or drawn keypoint
circles as in Figure~\ref{softmax-sum}.}
\label{arch}
\end{center}
\vskip -0.2in
\end{figure}

Our architecture follows \cite{siarohin2020order} without the dense
motion module, after changing its keypoint generation module to return our
mask. After the mask is obtained, we follow the encoder decoder approach as
\cite{shalev2020image}, which had success in similar tasks
\cite{newell2016stacked}.

Namely, the encoder of the low resolution generator consists of $\textit{conv}_{7
\times 7}$, $\textit{batch\_norm}$, $\textit{relu}$, followed by six
residual blocks of $\textit{batch\_norm}$, $\textit{relu}$,
$\textit{conv}_{3 \times 3}$ ,$\textit{batch\_norm}$, $\textit{relu}$,
$\textit{conv}_{3 \times 3}$, (and a sum with the source).
The residual blocks help to maintain the identity of the source image
\cite{he2015deep}.
The decoder consists of two blocks, each is a sequence of
$\textit{up\_sample}_{2 \times 2 }$, $\textit{batch\_norm}$,
$\textit{relu}$. The decoder is followed by a $\textit{conv}_{7 \times 7}$
and a $\textit{sigmoid}$ activation.
For the high resolution generator, use an encoder (decoder) with five encoding (decoding) blocks,
where each block is a sequence of
$\textit{conv}_{3 \times 3}$, $\textit{batch\_norm}$, $\textit{relu}$
$\textit{avg\_pool}_{2 \times 2}$, and each decoding block is a sequence of
$\textit{up\_sample}_{2 \times 2}$, $\textit{conv}_{3 \times 3}$,
$\textit{batch\_norm}$, $\textit{relu}$.
We add skip connections from each of the encoding layers to its
corresponding encoding layer, to form a U-Net architecture
\cite{ronneberger2015unet}.

\subsection{Losses}
We use the same perceptual loss as in \cite{siarohin2020order} which is
based on the implementation of \cite{wang2018videotovideo}. With the input
driving frame $D$ and the corresponding reconstructed frame $\hat{D}$, the
reconstruction loss is written as: $L_{rec}(\hat{D}, D) =
\sum_{i=1}^{I} |N_i(\hat{D})-N_i(D)|$, where $N_i(\cdot)$ is the $i^{th}$
channel feature extracted from a specific VGG-19 layer \cite{simonyan2015deep}
and $I$ is the number of feature channels in this layer. Additionally we
use this loss on
a number of resolutions, forming a pyramid obtained by down-sampling
$\hat{D}$ and $D$, similarly to MS-SSIM \cite{1292216}, \cite{tang2019dual}.
The resolutions are $256
\times 256$, $128 \times 128$, $64 \times 64$ and $32 \times 32$.

\section{Experiments}
\subsection{Datasets}
The Tai-chi-HD dataset, which includes brief videos of people doing Tai-chi
exercises, was used for training and evaluation. Following
\cite{siarohin2020order}, 3,141 Tai-chi videos were downloaded from YouTube.
The videos were cropped and resized to a resolution of $256^2$, while the
aspect ratio was preserved. There are 3,016 training videos and 125
evaluation videos.

\subsection{Comparison with Previous Works}
In order to compare our work to previous works (Table~\ref{table:results})
we used metrics previously used in similar papers.
Average Key-points Distance \cite{cao2017realtime} (AKD)
measures the average key-points distance between the generated video and
the source video. Average Euclidean Distance \cite{zheng2019joint} (AED)
measures the average euclidean distance
between the representations of the ground-truth and generated videos in
some embedding space. In addition, we added the L1 distance as well.
Our AED and AKD metrics were calculated using the following repository:
\url{https://github.com/AliaksandrSiarohin/pose-evaluation}
\\
Note that these metrics aren't optimal as one can easily improve
reconstruction by increasing the bottleneck, and we can see our
artifacts in animation results (Table~\ref{table:images}).
However, our approach did follow the structure of driving video well and
better than the others,
and to improve the identity and background artifacts,
we suggest some fixes in Section~\ref{future}.
Due to the smaller bottleneck of the second (softmax) mask, the pose
worsened, but the generalization for the background improved (which also
has to do with the low but non zero grayscale values given to the
background in the heatmap mask representation, which can be solved with
thresholding - see Section~\ref{future}).


\begin{table}[t]
\caption{Accuracy Metrics}
\label{table:results}
\vskip 0.15in
\begin{center}
\begin{small}
\begin{sc}
\begin{tabular}{lcccr}
\toprule
Method & AKD & AED & L1 \\
\midrule
X2Face    & 17.654 & 0.272 & 0.080 \\
Monkey-Net    & 10.798 & 0.228 & 0.077 \\
FOMM    & 6.872 & 0.167 & 0.063 \\
Perturbed Mask & 4.239 & 0.147 & 0.047 \\
Ours (circles mask) & 14.760& 0.245 & 0.077 \\
Ours & 5.551 & 0.141 &  0.045\\
\midrule
Improvement (FOMM)    & 19.2\% & 15.5\% & 28.5\% \\
\bottomrule
\end{tabular}
\end{sc}
\end{small}
\end{center}
\vskip -0.1in
\end{table}

\subsubsection{Qualitative comparison}
When comparing our work visually to \cite{siarohin2020order}, we can argue
that we obtained a better pose which is almost the same with the original.
We can observe a closer synthesized leg in the second frame (to the
original driving), and a better pose (specifically the hands) on the third and
forth frames.
However, the background generated is damaged, due to the lack of
thresholding in our mask which means we will always have noise
to the background generation.
See Section~\ref{future} for future suggestions.
We can also see that the softmax (circular keypoint mask) version worked
fine, preserved the background, but with a worse pose, due to less
information in this type of mask.
\begin{table}[t]
\caption{Images comparison}
\label{table:images}
\vskip 0.15in
\begin{center}
\begin{small}
\begin{sc}
\begin{tabular}{m{1.0cm}m{1.0cm}m{1.0cm}m{1.0cm}m{1.0cm}m{1.0cm}}
\toprule
Source image & Driving\\
\toprule
\includegraphics[width=1cm, height=1cm]{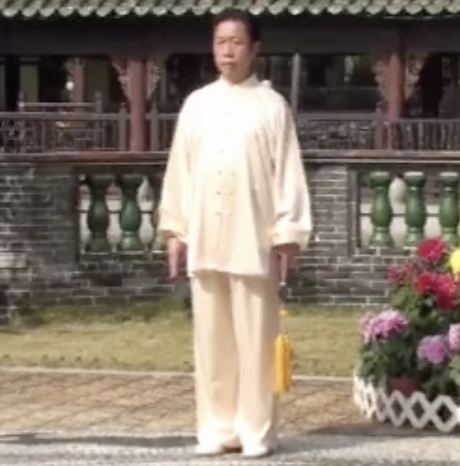} &
\includegraphics[width=1cm, height=1cm]{images/intro_image/Driving_1.png} &
\includegraphics[width=1cm, height=1cm]{images/intro_image/Driving_2.png} &
\includegraphics[width=1cm, height=1cm]{images/intro_image/Driving_4.png} &
\includegraphics[width=1cm, height=1cm]{images/intro_image/Driving_5.png} &
\includegraphics[width=1cm, height=1cm]{images/intro_image/Driving_6.png} \\
\midrule
FOMM & \includegraphics[width=1cm, height=1cm]{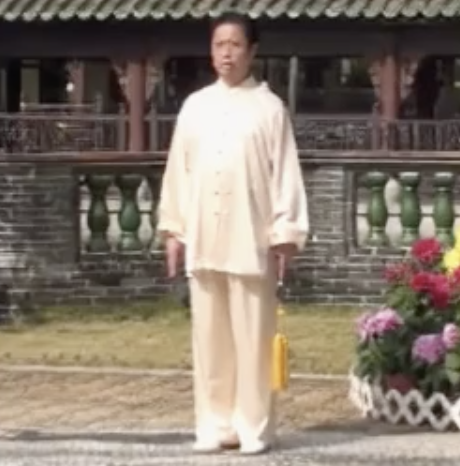} &
\includegraphics[width=1cm, height=1cm]{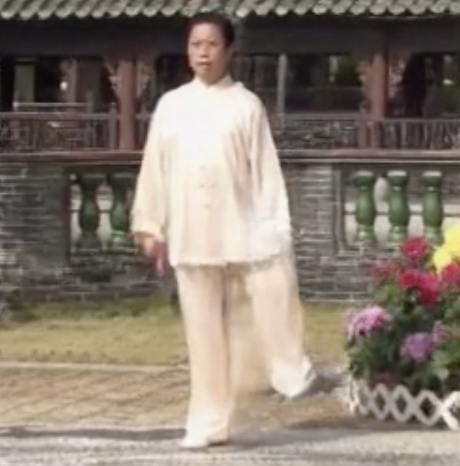} &
\includegraphics[width=1cm, height=1cm]{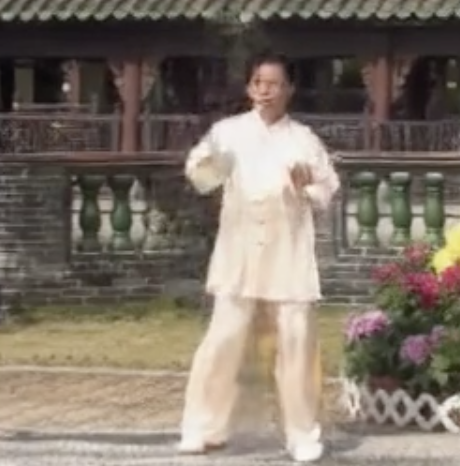} &
\includegraphics[width=1cm, height=1cm]{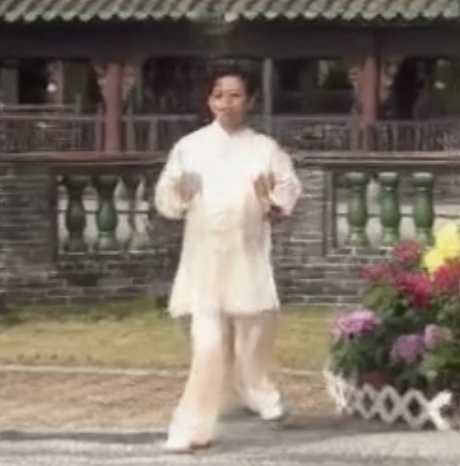} &
\includegraphics[width=1cm, height=1cm]{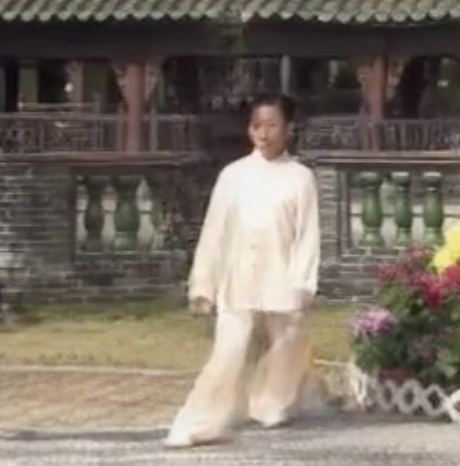} \\
Ours (circles)&
\includegraphics[width=1cm, height=1cm]{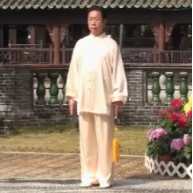} &
\includegraphics[width=1cm, height=1cm]{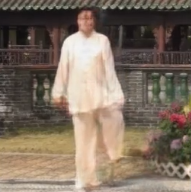} &
\includegraphics[width=1cm, height=1cm]{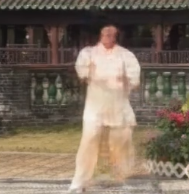} &
\includegraphics[width=1cm, height=1cm]{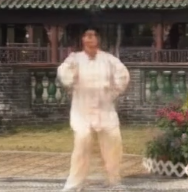} &
\includegraphics[width=1cm, height=1cm]{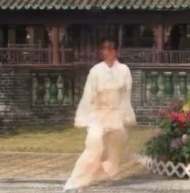} \\

Ours (mask) & \includegraphics[width=1cm, height=1cm]{images/intro_image/animate_1.png} &
\includegraphics[width=1cm, height=1cm]{images/intro_image/animate_2.png} &
\includegraphics[width=1cm, height=1cm]{images/intro_image/animate_4.png} &
\includegraphics[width=1cm, height=1cm]{images/intro_image/animate_5.png} &
\includegraphics[width=1cm, height=1cm]{images/intro_image/animate_6.png} \\
\bottomrule
\end{tabular}
\end{sc}
\end{small}
\end{center}
\vskip -0.1in
\end{table}

\section{Future work}
\label{future}
We would like to test our module on more datasets, and compare them to the
state of the art. In addition, summing the heatmap channels might not be
optimal, and there is certainly some space to try something deeper with the
features extracted in the keypoint detector as an input, or feed all of the
channels separately into the generator. We want to experiment with
mask thresholds due to the distortions in the animation background,
similarly to \cite{shalev2020image}.
We may also increase the number of keypoints, but that would probably be
more beneficent to the second type of mask, and would increase the
required GPU memory proportionally. We also suggest (in the second, circles
only mask) coloring
matching keypoints with the same color to help the module to learn a motion
flow.

\section{Conclusions}
We constructed a novel method for image animation by moving the need for
a strong motion prior (optical flow) to the assumption of a pre-trained
keypoint detector/keypoint heatmaps prior to activation, which might be
based on a motion prior.
By doing so, we encapsulated motion to a motion mask, which is
bottlenecked by the prior training which has the keypoint bottleneck.
The motion masks are then fed into a generator, which combines the
appearance of the source image and the mask which represents the structure,
decoupled from any appearance naturally by the assumption that during the
training of the keypoint detector, the heatmap mask went into a keypoint
bottleneck. After evaluation, we can conclude that our method is
feasible and improves the pose although with some artifacts.
\section*{Software Data and more result videos}
Detailed in our repository:
\\
\url{https://github.com/or-toledano/
animation-with-keypoint-mask}
\bibliography{animation-with-keypoint-mask}
\bibliographystyle{icml2021}

\end{document}